\documentclass{article}
\usepackage{amsmath}
\usepackage{amsfonts}
\usepackage{dsfont}
\usepackage{tikz}
\usepackage{subfig}

\graphicspath{{assets/}}

\usepackage{microtype}
\usepackage{graphicx}
\usepackage{booktabs} 

\usepackage{hyperref}

\usepackage[accepted]{src/icml2020}

\icmltitlerunning{Hierarchical reinforcement learning for exploration and transfer}

\begin{document}

\twocolumn[
\icmltitle{Hierarchical reinforcement learning for efficient exploration and transfer}




\begin{icmlauthorlist}
\icmlauthor{Lorenzo Steccanella}{upf}
\icmlauthor{Simone Totaro}{upf}
\icmlauthor{Damien Allonsius}{upf}
\icmlauthor{Anders Jonsson}{upf}
\end{icmlauthorlist}

\icmlaffiliation{upf}{DTIC, Universitat Pompeu Fabra, Barcelona, Spain}

\icmlcorrespondingauthor{Lorenzo Steccanella}{lorenzo.steccanella@upf.edu}

\icmlkeywords{Hierarchical Reinforcement Learning, Reinforcement Learning, Lifelong, Exploration, ICML}

\vskip 0.3in
]



\printAffiliationsAndNotice{}  

\begin{abstract}
Sparse-reward domains are challenging for reinforcement learning algorithms since significant exploration is needed before encountering reward for the first time. Hierarchical reinforcement learning can facilitate exploration by reducing the number of decisions necessary before obtaining a reward. In this paper, we present a novel hierarchical reinforcement learning framework based on the compression of an invariant state space that is common to a range of tasks. The algorithm introduces subtasks which consist in moving between the state partitions induced by the compression. Results indicate that the algorithm can successfully solve complex sparse-reward domains, and transfer knowledge to solve new, previously unseen tasks more quickly.
\end{abstract}

\section{Introduction}\label{sec: introduction}



    


In reinforcement learning, an agent attempts to maximize its cumulative reward through interaction with an unknown environment. In each round, the agent observes a state, takes an action, receives an immediate reward, and transitions to a next state. The aim of the agent is to learn a policy, i.e.~a mapping from states to actions, that maximizes the expected future sum of rewards. To do so, the agent has to explore the environment by taking actions and observing their effects, and exploit its current knowledge by repeating action choices that have been successful in the past.

An important challenge in reinforcement learning is solving domains with sparse rewards, i.e.~when the immediate reward signal is almost always zero. In this case, all actions initially appear equally good, and it becomes important to explore efficiently until the agent finds a high-reward state. Only then does it become possible to distinguish actions that eventually lead to high reward.

Hierarchical reinforcement learning (HRL) exploits structure in the environment to decompose complex tasks into simpler subtasks \cite{dayan1993feudal,sutton1999between,dietterich2000hierarchical}. HRL provides a mechanism for acting on different timescales by introducing temporally extended actions that solve the subtasks, and can help alleviate the problem of exploration in sparse-reward domains since temporally extended actions reduce the number of decisions necessary to reach high-reward states.

Early work on HRL showed that it is important to exploit structure both in time and space, i.e.~for the decomposition to accelerate learning, the subtasks have to be significantly easier to solve than the original task. This is usually made possible by a compressed state space in the form of state abstraction \cite{dietterich2000hierarchical}. When the state space is sufficiently compressed, state-based methods can be used to solve the subtasks, significantly outperforming non-hierarchical methods in many cases. However, even compressed state spaces become large in complex tasks.

    On the other hand, HRL methods using subgoals to guide exploration, either as part of the value function representation \cite{nachum2018near,schaul2015universal,sutton2017horde}, or as pseudo-reward \cite{eysenbach2019search,florensa2017automatic}, have shown progress in hard exploration tasks, even for high-dimensional state and action spaces. These methods are not as sample efficient, however.

    In this paper we propose a novel hierarchical reinforcement learning framework that attempts to exploit the best of both worlds. We use a fixed, state-dependent compression function to define a hierarchical decomposition of complex, sparse-reward tasks. The agent defines subtasks which consist in navigating across state-space partitions by jointly learning the policy of each temporally extended action. The compression function makes it possible to use tabular methods at the top level to effectively explore large state spaces even in sparse reward settings. Furthermore we show that our method is suitable for transfer learning across tasks that are defined by introducing additional learning components.

\section{Background}\label{sec: background}

In this section we define several concepts and associated notation used throughout the paper.

\subsection{Markov Decision Process}

		We consider a Markov Decision Process (MDP) \cite{puterman2014markov} defined by the tuple $\mathcal{M} = \langle S,A,r,P \rangle$, where $S$ is the finite state space, $A$ is the finite action space, $r:S \times A \rightarrow \mathbb{R}$ is the Markovian reward function, and $P: S \times A \rightarrow \Delta(S) $ is the transition kernel. Here, $\Delta(S)$ is the probability simplex on $S$, i.e.~the set of all probability distributions over $S$. At time $t$, the agent observes state $s_t \in S$, takes an action $a_t \in A$, obtains reward $r_t$ with expected value $\mathbb{E}[r_t] = r(s_t, a_t)$, and transitions to a new state $s_{t+1} \sim P(\cdot \rvert s_t, a_t)$. We refer to $(s_t,a_t,r_t,s_{t+1})$ as a {\em transition}.

		Let $\pi$ denote a stochastic policy $\pi : S \rightarrow \Delta(A)$ and $\eta(\pi)$ its expected discounted cumulative reward under some initial distribution $d_0\in\Delta(S)$ over states: 
\[
			\eta(\pi) = E_{s\sim d_0} [V^{\pi}(s)].
\]
Here, $V^{\pi}(s)$ is the value function of policy $\pi$ in state $s$,
\[
				V^{\pi}(s) = \mathbb{E} \left[ \left. \sum_{t=1}^{\infty} \gamma^{t-1} r(s_t,a_t) \right\vert s_1=s \right],
\]
			 where $\gamma\in(0,1]$ is a discount factor and the expectation is over $P$ and $\pi$.
			We also define the action-value function $Q^\pi$ of $\pi$ in state-action pair $(s,a)$ as
\[
				Q^{\pi}(s, a) = \mathbb{E} \left[ \left. \sum_{t=1}^{\infty} \gamma^{t-1} r(s_t,a_t) \right\vert s_1=s, a_1=a \right].
\]
			The goal of the agent is to find a policy that maximizes the expected discounted cumulative reward $\eta(\pi)$:
\[
			\pi^{*}     = \arg \max_{\pi} \eta(\pi).
\]

\subsection{Options Framework}
	Given an MDP $\mathcal{M} = \langle S,A,r,P \rangle$, an option is a temporally extended action $o=\langle I^o, \pi^o, \beta^o \rangle$, where $I^o \subseteq S$ is an initiation set, $\pi^o: S \rightarrow \Delta(A)$ is a policy and $\beta^o: S \rightarrow [0, 1]$ is a termination function \cite{sutton1999between}. Adding options to the action set $A$ of $\mathcal{M}$ forms a Semi-Markov Decision Process (SMDP), which enables an agent to act and reason on multiple timescales. To train the policy $\pi^o$, it is common to define an option-specific reward function $r^o$.

If an option $o$ is selected in state $s_t\in I^o$ at time $t$, the option takes actions using policy $\pi^o$ until it reaches a state $s_{t+k}$ in which the termination condition $\beta^o(s_{t+k})$ triggers. Although the option takes multiple actions, from the perspective of the SMDP a single decision takes place at time $t$, and the reward accumulated until time $t+k$ is $r_t + \gamma r_{t+1} + \ldots + \gamma^{k-1} r_{t+k-1}$. The theory of MDPs extends to SMDPs, e.g.~we can define a policy $\pi:S\rightarrow\Delta(O)$ over a set of options $O$, a value function $V^\pi$ of this policy and an action-value function $Q^\pi$ over state-option pairs.

\section{Algorithm}\label{sec: setting}

In this section we describe our algorithm for constructing an SMDP that can solve a range of different tasks.

\subsection{Task MDPs}\label{sec: a_mdp}

		We assume that each task $\mathcal{T}$ is described by an MDP $\mathcal{M}_\mathcal{T} = \langle S_i\times S_{\mathcal{T}},A_i\cup A_{\mathcal{T}},r_{\mathcal{T}},P_i\cup P_{\mathcal{T}} \rangle$. Crucially, the state-action space $S_i\times A_i$ as well as the transition kernel $P_i:S_i\times A_i\rightarrow \Delta(S_i)$ are {\em invariant}, i.e.~shared among all tasks. On the other hand, the state-action space $S_{\mathcal{T}}\times A_{\mathcal{T}}$, reward function $r_{\mathcal{T}}:S_{\mathcal{T}}\times A_{\mathcal{T}}\rightarrow\mathbb{R}$ and transition kernel $(S_i\cup S_{\mathcal{T}}) \times A_{\mathcal{T}}\rightarrow\Delta(S_{\mathcal{T}})$ are task-specific. We assume that actions in $A_i$ incur zero reward, and that states in $S_i$ are unaffected by actions in $A_{\mathcal{T}}$. Task $\mathcal{T}$ is only coupled to the invariant MDP through the transition kernel $P_{\mathcal{T}}$, since the effect of actions in $A_{\mathcal{T}}$ depend on the states in $S_i$.

\subsection{Invariant SMDP}

		We further assume that the agent has access to a partition $Z=\{Z_1,\ldots,Z_m\}$ of the invariant state space, i.e.~$S_i=Z_1\cup\cdots\cup Z_m$ and $Z_i\cap Z_j=\emptyset$ for each pair $(Z_i,Z_j)\in Z^2$. Even though each element of $Z$ is a subset of $S_i$, we often use lower-case letters to denote elements of $Z$, and we refer to each element $z\in Z$ as a {\em region}. We use the partition $Z$ to form an SMDP over the invariant part of the state-action space. This SMDP is defined as $\mathcal{S} = \langle Z, O, P_Z \rangle$, where $Z$ is the set of regions, $O$ is a set of options, and $P_Z:Z\times O\rightarrow \Delta(Z)$ is a transition kernel.

We first define the set of neighbors of a region $z\in Z$ as
\[
\mathcal{N}(z) = \{z': \exists (s,a,s') \in z\times A_i\times z', P_i(s'|s,a) > 0\}.
\]
Hence neighbors of $z$ can be reached in one step from some state in $z$. For each neighbor $z'\in\mathcal{N}(z)$, we define an option $o_{z,z'}=\langle z, \pi_{z,z'}, \beta_z \rangle$ whose subtask is to reach region $z'$ from $z$. Hence the initiation set is $z$, the termination function is $\beta_z(s)=0$ if $s\in z$ and $\beta_z(s)=1$ otherwise, and the policy $\pi_{z,z'}$ should reach region $z'$ as quickly as possible. 

The set of options available to the agent in region $z\in Z$ is $O_z=\{o_{z,z'} : z' \in \mathcal{N}(z)\} \subseteq O$, i.e.~all options that can be initiated in $z$ and that transition to a neighbor of $z$. Note that the option sets $O_z$ are disjoint, i.e.~each region $z$ has its own set of admissible options. The transition kernel $P_Z$ determines how successful the options are; ideally, $P_Z(z' | z, o_{z,z'})$ should be close to $1$ for each pair of neighboring regions $(z,z')$, but can be smaller to reflect that $o_{z,z'}$ sometimes ends up in a region different from $z'$.
			
\subsection{Option MDPs}\label{sec: partial_mdps}

	We do not assume that the policy $\pi_{z,z'}$ of each option $o_{z,z'}$ is given; rather, the agent has to learn the policy $\pi_{z,z'}$ from experience. For this purpose, we define an option-specific MDP $M_{z,z'}=\langle S_z, A_i, P_z, r_{z,z'} \rangle$ associated with option $o_{z,z'}$. Here, the state space $S_z=z\cup\mathcal{N}(z)$ consists of all states in the region $z$, plus all the neighboring regions of $z$. The set of actions $A_i$ are those of the invariant part of the state-action space. All the states in $\mathcal{N}(z)$ are terminal states. The transition kernel $P_z$ is a projection of the invariant transition kernel $P_i$ onto the state-action space $z\times A_i$ involving non-terminal states, and is defined as
\[
P_z(s'|s,a) =
\left\{
\begin{array}{ll}
P_i(s'|s,a), & \mathrm{if} \; s'\in z,\\
\sum_{s''\in s'} P_i(s''|s,a) & \mathrm{if} \; s'\in \mathcal{N}(z).
\end{array}
\right.
\]
Hence the probability of transitioning to a neighbor $s'$ of $z$ is the sum of probabilities of transitioning to any state in $s'$.

In the definition of $M_{z,z'}$, the state-action space $S_z\times A_i$ and transition kernel $P_z$ are shared among all options in $O_z$. They only differ in the reward function $r_{z,z'}: z\times A_i\times S_z\rightarrow \mathbb{R}$ defined on triples $(s,a,s')$, i.e.~it depends on the resulting next state $s'$. The theory of MDPs easily extends to this case. Specifically, the reward function $r_{z,z'}$ is defined as
\begin{align}\label{eqn: partial_reward}
r_{z,z'}(s,a,s') &=
\left\{
\begin{array}{rl}
+0.8, & \mathrm{if} \; s' = z',\\
-0.1, & \mathrm{if} \; s' \in \mathcal{N}(z) \setminus \{z'\}.
\end{array}
\right.
\end{align}
In other words, successfully terminating in region $z'$ is awarded with a reward of $+0.8$, while terminating in a region different from $z'$ is penalized with a reward of $-0.1$. If the option can't terminate in a time limit of 100 steps the same negative reward $-0.1$ is given. In practice, option $o_{z,z'}$ can compute the policy $\pi_{z,z'}$ indirectly by maintaining a value function $V_{z,z'}$ associated to the option MDP $M_{z,z'}$.
	
\subsection{Algorithm}\label{sec: algorithm}

	In practice, we do not assume that the agent has access to the invariant SMDP $\mathcal{S} = \langle Z, O, P_Z \rangle$. Instead, the agent can only observe the current state $s\in S_i$, select an action $a\in A_i$, and observe the next state $s' \sim P_i(\cdot | s,a)$. Rather than observing regions in $Z$, the agent has oracle access to a compression function $f:S_i \rightarrow \mathbb{N}_+$ from invariant states to non-negative integers. Each region $z$ has an associated integer ID $N(z)$ and is implicitly defined as $z=\{s\in S_i : f(s) = N(z)\}$. To identify regions, the agent has to repeatedly query the function $f$ on observed states and store the integers returned. By abuse of notation we often use $z$ to denote both a region in $Z$ and its associated ID.

Our algorithm iteratively grows an estimate of the invariant SMDP $\mathcal{S} = \langle Z, O, P_Z \rangle$. Initially, the agent only observes a single state $s\in S_i$ and associated region $z=f(s)$. Hence the state space $Z$ contains a single region $z$, whose associated option set $O_z$ is initially empty. In this case, the only alternative available to the agent is to {\em explore}. For each region $z$, we add an exploration option $o_z^e=\langle z, \pi_z^e, \beta_z \rangle$ to the option set $O$. This option has the same initiation set and termination condition as the options in $O_z$, but the policy $\pi_z^e$ is an exploration policy that selects actions at random or implements a more advanced exploration strategy.

\begin{figure}
  \centering
  \begin{tikzpicture}
    \def\k{0.3}
    \def\e{1.7}
    
    \draw (0,0) circle (\k cm);
    \draw (\e,0) circle (\k cm);
    \draw (0,\e) circle (\k cm);
    \draw (-\e,0) circle (\k cm);
    \draw (0,-\e) circle (\k cm);

    \draw[very thick, ->] (\k, 0) -- (\e - \k, 0);
    \draw[very thick, ->] (0, -\k) -- (0, -\e + \k);
    \draw[very thick, ->] (0, \k) -- (0, \e-\k);
    \draw[very thick, ->] (\e, \k) -- (\k, \e);
    \draw[very thick, ->] (-\k, \e) -- (-\e, \k);
    \draw[very thick, ->] (-\e, -\k) -- (-\k, -\e);
    \draw[very thick, ->] (\k, -\e) -- (\e, -\k);

    \node at (0,0) {\color{red} {$z_1$}};
    \node at (\e,0) {$z_2$};
    \node at (0,\e) {$z_3$};
    \node at (-\e,0) {$z_4$};
    \node at (0,-\e) {$z_5$};

    \node[left] at (0,\e/2) {$o_{z_1, z_3}$};
    \node[below] at (\e/2,0) {$o_{z_1, z_2}$};
    \node[left] at (0, -\e / 2) {$o_{z_1, z_5}$};
    \node[right] at (\e/1.4,-\e/2) {$o_{z_5, z_2}$};
    \node[right] at (\e/1.4,\e/2) {$o_{z_2, z_3}$};
    \node[left] at (-\e/1.4,\e/2) {$o_ {z_3, z_4}$};
    \node[left] at (-\e/1.4,-\e/2) {$o_{z_4, z_5}$};

  \end{tikzpicture}
  \caption{Example representation discovered by the algorithm.}
  \label{fig:graph_manager}
\end{figure}
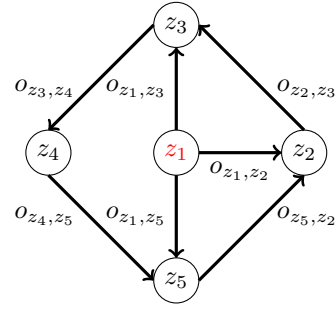

Once the agent discovers a neighboring region $z'$ of $z$, it adds region $z'$ to the set $Z$ and the associated option $o_{z,z'}$ to the option set $O$. The agent also maintains and updates a directed graph whose nodes are regions and whose edges represent the neighbor relation. Hence next time the agent visits region $z$, one of its available actions is to select option $o_{z,z'}$. When option $o_{z,z'}$ is selected, it chooses actions using its policy $\pi_{z,z'}$ and simultaneously updates $\pi_{z,z'}$ based on the rewards of the option MDP $M_{z,z'}$. Figure~\ref{fig:graph_manager} shows an example representation discovered by the algorithm.

Algorithm \ref{algo: hrl} shows pseudo-code of the algorithm. As explained, $Z$ is initialized with the region $z$ of the initial state $s$, and $O$ is initialized with the exploration option $o_z^e$. In each iteration, the algorithm selects an option $o$ which is applicable in the current region $z$. This option then runs from the current state $s$ until terminating in a state $s'$ whose associated region $z'$ is different from $z$. If this is the first time region $z'$ has been observed, it is added to $Z$ and the exploration option $o_{z'}^e$ is appended to $O$. If this is the first time $z'$ has been reached from $z$, the option $o_{z,z'}$ is appended to $O$. The process then repeats from state $s'$ and region $z'$.

The subroutine $\textsc{GetOption}$ that selects an option $o$ in the current region $z$ can be implemented in different ways. If the aim is just to estimate the invariant SMDP $\mathcal{S} = \langle Z, O, P_Z \rangle$, the optimal choice of option is that which maximizes the chance of discovering new regions or, alternatively, that which improves the ability of options to successfully solve their subtasks. If the aim is to solve a task $\mathcal{T}$, the optimal choice of option is that which maximizes the reward of $\mathcal{T}$. On the other hand, the subroutine $\textsc{RunOption}$ executes the policy of the option while simultaneously improving the associated option policy.

			\begin{algorithm}
			\caption{\textsc{InvariantHRL}}
			\begin{algorithmic}[1]
				 \STATE {\bf Input}: Action set $A_i$, oracle compression function $f$
				 \STATE $s \gets $ initial state, $z \gets f(s)$
				 \STATE $Z \gets \{z\}$, $O \gets \{o_z^e\}$

				 \WHILE{within budget}
					 \STATE $o  \gets \textsc{GetOption}(z,O)$
					 \STATE $s' \gets \textsc{RunOption}(s,o,A_i)$, $z' \gets f(s')$
					 \IF{$z' \notin Z$}
						\STATE $Z \gets Z \cup \{z'\}$
						\STATE $O \gets O \cup \{o_{z'}^e\}$
					 \ENDIF
					 \IF{$o_{z,z'} \notin O$}
						\STATE $O \gets O \cup \{o_{z,z'}\}$
					 \ENDIF
					 \STATE $s \gets s'$, $z \gets z'$
				\ENDWHILE
			 \end{algorithmic}
			 \label{algo: hrl}
			\end{algorithm}

\subsection{Properties}

	The proposed algorithm has several advantages. Both the invariant SMDP $\mathcal{S}$ and the option MDPs $M_{z,z'}$ have much smaller state-action spaces than $S_i\times A_i$, which leads to faster learning. In addition, on the SMDP level, distant regions are reached by relatively few decisions, which facilitates exploration. Even if the state space $S_i$ is high-dimensional, the number of regions is relatively small, which makes it possible to store region-specific information. Once learned, the estimate of the invariant SMDP $\mathcal{S}$ can be reused in many tasks, which facilitates transfer.

	The main drawback of the algorithm is that the number of options grows with the size of the region set $Z$, each requiring the solution of an additional option MDP.

\subsection{Solving tasks}

Recall that each task $\mathcal{T}$ is defined by a task MDP $\mathcal{M}_\mathcal{T} = \langle S_i\times S_{\mathcal{T}},A_i\cup A_{\mathcal{T}},r_{\mathcal{T}},P_i\cup P_{\mathcal{T}} \rangle$. Given an estimate $\mathcal{S}=\langle Z,O,P_Z\rangle$, we define an associated task SMDP $\mathcal{S}_{\mathcal{T}} = \langle Z_{\mathcal{T}}, O\cup O_{\mathcal{T}}, r_{\mathcal{T}}, P_Z\cup P_{\mathcal{T}}'  \rangle$. Here, $O_{\mathcal{T}}$ is a set of task-specific options whose purpose is to change the task state in $S_{\mathcal{T}}$, and $P_{\mathcal{T}}'$ is the transition kernel corresponding to these options. The state space is $Z_{\mathcal{T}} = Z \times S_{\mathcal{T}}$, i.e.~a state $(z,s)\in Z_{\mathcal{T}}$ consists of a region $z$ and a task state $s$.

As before, we do not assume that the agent has access to options in $O_{\mathcal{T}}$. Instead, the agent has to discover from experience how to change the task state in $S_{\mathcal{T}}$. For this purpose, we redefine the exploration option $o_z^e$ of each region $z$ so that it has access to actions in $A_{\mathcal{T}}$. When selected in state $(z,s)$, $o_z^e$ may terminate for one of two reasons: either the current region changes, i.e.~the next state is $(z',s)$ for some neighbor $z'$ of $z$, or the current task state changes, i.e.~the next state is $(z,s')$ for some task state $s'$. In the latter case, the agent will add an option $o_z^{s,s'}$ to $O_{\mathcal{T}}$ which is applicable in $(z,s)$ and whose subtask is to reach state $(z,s')$. Option $o_z^{s,s'}$ has an associated option MDP $M_z^{s,s'}$, analogous to $M_{z,z'}$ except that it assigns positive reward to $(z,s')$.

To solve task $\mathcal{T}$, the agent need to maintain and update a high-level policy $\pi_{\mathcal{T}}: Z_{\mathcal{T}} \rightarrow \Delta(O\cup O_{\mathcal{T}})$ for the task SMDP $\mathcal{S}_{\mathcal{T}}$. In a state $(z,s)$, policy $\pi_{\mathcal{T}}$ has to decide whether to change regions by selecting an option in $O$, or to change task states by selecting an option in $O_{\mathcal{T}}$. Because of our previous assumption on the reward $r_{\mathcal{T}}$, only options in $O_{\mathcal{T}}$ will incur non-zero reward, which has to be appropriately discounted after applying each option. Note that in Algorithm 1, policy $\pi_{\mathcal{T}}$ plays the role of the subroutine $\texttt{GetOption}$.

The transition kernel $P_{\mathcal{T}}'$ measures the ability of task options in $O_{\mathcal{T}}$ to successfully solve their subtasks. Hence $P_{\mathcal{T}}'((z,s')|(z,s),o_z^{s,s'})$ should be close to $1$, but is lower in case option $o_z^{s,s'}$ sometimes terminates in the wrong state. In our experiments, however, the agent performs model-free learning and never estimates the transition kernel $P_{\mathcal{T}}'$. 

\subsection{Controllability} \label{sec: credit}

According to the definition of the option reward function $r_{z,z'}$ in \eqref{eqn: partial_reward}, option $o_{z,z'}$ is equally rewarded for reaching any boundary state between regions $z$ and $z'$. However, all boundary states may not be equally valuable, i.e.~from some boundary states the options in $O_{z'}$ may have a higher chance of terminating successfully. To encourage option $o_{z,z'}$ to reach valuable boundary states and thus make the algorithm more robust to the choice of compression function $f$, we add a reward bonus when the option successfully terminates in a state $s'$ belonging to region $z'$.

One possibility is that the reward bonus depends on the value of state $s'$ of options in the set $O_{z'}$. However, this introduces a strong coupling between options in the set $O$: the value function $V_{z,z'}$ of option $o_{z,z'}$ will depend on the value functions of options in $O_{z'}$, which in turn depend on the value functions of options in neighboring regions of $z'$, etc. We want to avoid such a strong coupling since learning the option value functions may become as hard as learning a value function for the original invariant state space $S_i$.

Instead, we introduce a reward bonus which is a proxy for controllability, by counting the number of successful applications of subsequent options after $o_{z,z'}$ terminates. Let $M$ be the number of options that are selected after $o_{z,z'}$, and let $N\leq M$ be the number of such options that terminate successfully. We define a controllability coefficient $\rho$ as
\begin{equation}
	\rho(z) = \frac{N}{M}.
\end{equation}
We then define a modified reward function $\bar{r}_{z,z'}$ which equals $r_{z,z'}$ except when $o_{z,z'}$ terminates successfully, i.e.~$\bar{r}_{z,z'}(s,a,s') = r_{z,z'}(s,a,s') + \rho(z)$ if $s'\in z'$. In experiments we use a fixed horizon $M=10$ after which we consider successful options transitions as not relevant. In practice, the algorithm has to wait for 10 more options before assigning reward to the last transition of option $o_{z,z'}$.

\section{Implementation}\label{sec: implementation}

In this section we describe the implementation of our algorithm. We distinguish between a {\em manager} in charge of solving the task SMDP $\mathcal{S}_{\mathcal{T}}$, and {\em workers} in charge of solving the option MDPs $M_{z,z'}$ (or $M_z^{s,s'}$ for task options).

\subsection{Manager} \label{sec: manager}
	Since the space of regions $Z$ is small, the manager performs tabular Q-learning over the task SMDP $\mathcal{S}_{\mathcal{T}}$. This procedure is shown in Algorithm~\ref{algo: manager}. Similar to Algorithm 1, the task state space $S_{\mathcal{T}}$ and option set $O_{\mathcal{T}}$ are successively grown as the agent discovers new states and transitions.

			\begin{algorithm}
			\caption{\textsc{Manager}}
			\begin{algorithmic}[1]
				 \STATE {\bf Input}: Task action set $A_{\mathcal{T}}$, invariant SMDP $\mathcal{S}$
				 \STATE $z \gets $ initial region, $s \gets $ initial task state
				 \STATE $S_{\mathcal{T}} \gets \{s\}$, $O_{\mathcal{T}} \gets \emptyset$
				 \STATE $\pi_{\mathcal{T}} \gets $ initial policy
				 \WHILE{within budget}
					 \STATE $o  \gets \textsc{GetOption}(\pi_{\mathcal{T}},(z,s),O\cup O_{\mathcal{T}})$
					 \STATE $(z',s'),r \gets \textsc{RunOption}((z,s),o,A_i\cup A_{\mathcal{T}})$
					 \STATE $\textsc{UpdatePolicy}(\pi_{\mathcal{T}}, (z,s), o, r, (z',s'))$
					 \IF{$s' \notin S_{\mathcal{T}}$}
						\STATE $S_{\mathcal{T}} \gets S_{\mathcal{T}} \cup \{s'\}$
					 \ENDIF
					 \IF{$o_z^{s,s'} \notin O_{\mathcal{T}}$}
						\STATE $O_{\mathcal{T}} \gets O_{\mathcal{T}} \cup \{o_z^{s,s'}\}$
					 \ENDIF
					 \STATE $(z,s) \gets (z',s')$
				\ENDWHILE
			 \end{algorithmic}
			 \label{algo: manager}
			\end{algorithm}

\subsection{Worker}\label{sec: worker}
		 The worker associated with option $o_{z,z'} \in O $ (resp.~$o_z^{s,s'}\in O_{\mathcal{T}}$) should learn a policy $\pi_{z,z'}$ (resp.~$\pi_z^{s,s'}$) that allows the manager to transition between two abstract states $z, z'$ (resp.~task states $s,s'$). We use Self-Imitation Learning (SIL) \cite{oh2018self} which benefits from an exploration bonus coming from the self-imitation component of the loss function. Moreover, since the critic update is off-policy, one can relabel failed transitions in order to speed up learning of the correct option behavior, similar to Hindsight Experience Replay \cite{andrychowicz2017hindsight}.

		 The architecture is made of two separate neural networks, one for the policy $\pi_{z,z'}^\theta$, parameterized on $\theta$, and one for the value function $V_{z,z'}^\psi$, parameterized on $\psi$. The agent minimizes the loss in \eqref{eqn: total_loss} via mini batch stochastic gradient descent, with on-policy samples:
		\begin{gather}
			\label{eqn: total_loss}
			L(\theta, \psi) = L(\hat{\eta}_\theta) + \alpha H^\pi + L(\hat{V}_\psi).
		\end{gather}

\section{Experiments}\label{sec: experiments}

\begin{figure}
    \centering
    \subfloat[]{{\includegraphics[width=2.9cm]{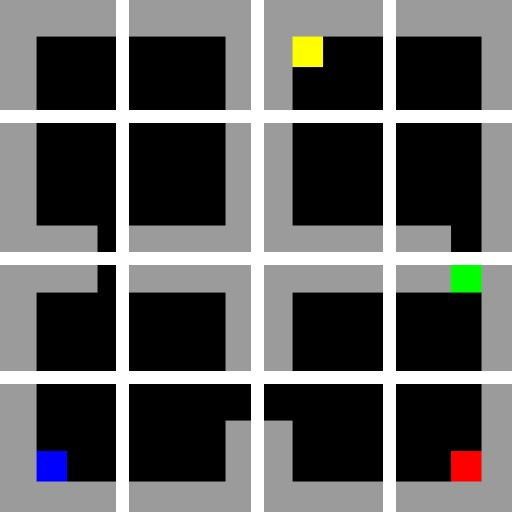} }}%
    \qquad
    \subfloat[]{{\includegraphics[width=4.4cm]{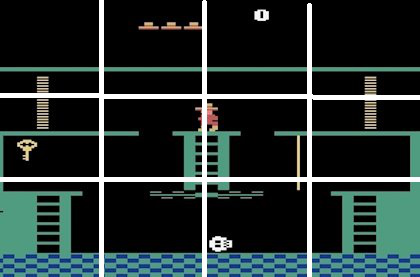} }}%
    \caption{Key-door-treasure-1 (a) and Montezuma's Revenge (b) with compression function superimposed.}%
    \label{fig:compression function}%
\end{figure}

To evaluate the proposed algorithm we use two benchmark domains: a Key-door-treasure GridWorld, and a simplified version of Montezuma's Revenge where the agent only has to pick up the key in the first room. In both domains, the invariant part of the state consists of the agent's location, and the compression function $f$ imposes a grid structure on top of the location (cf.~Figure~\ref{fig:compression function}). Results are averaged over 5 seeds and each experiment is run for 4e5 all the agents have been trained with the choice of hyperparameters in Figure \ref{fig: Hyperparameters}.

In the Key-door-treasure domain we make the reward progressively more sparse. In the simplest setting the agent obtains reward in each intermediate goal state, while in the hardest setting the agent obtains reward only in the terminal state. We also tested the transfer learning ability of our algorithm in new tasks generated by moving the position of the Key, Door and Treasure objects.

In Montezuma's Revenge, we evaluate whether our controllability proxy helps transition between regions. Montezuma does present an ideal environment to test this since imposing a grid set of regions on it does not respect the structural semantics of the environment and transitioning to the wrong state in another region may cause the agent to fall and die.

	Key-door-treasure is a stochastic variant of the original domain \cite{oh2018self} taking random actions with probability 20\%. The agent has a budget of 300 time steps.  We define two variants 
	and randomly generate multiple tasks by changing the location of the Key, Door and Treasure. In Key-door-treasure-1 \cite{oh2018self} the key is in the same room as the door, while in Key-door-treasure-2 the key is in a different room, making exploration harder.

\subsection{Exploration}\label{sec: exploration}

To investigate the exploration advantage of the proposed algorithm, we compare it against SIL \cite{oh2018self} and against a version of SIL augmented with count-based exploration \cite{strehl2008analysis} that gives an exploration bonus reward $r_{e x p}(s,a)=\beta / \sqrt{N(s)},$ where $N(s)$ is the visit count of state $s$ and $\beta$ is a hyperparameter. In the figures, our algorithm is labelled HRL-SIL, while SIL and SIL-EXP refer to SIL without/with the exploration bonus.

\begin{figure}
    \centering
    \subfloat[Reward for all objects. ]{{\includegraphics[width=3.65cm]{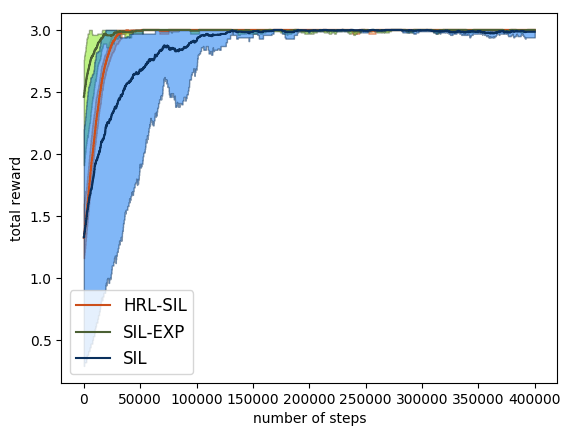} }}%
    \qquad
    \subfloat[Reward for treasure only.]{{\includegraphics[width=3.65cm]{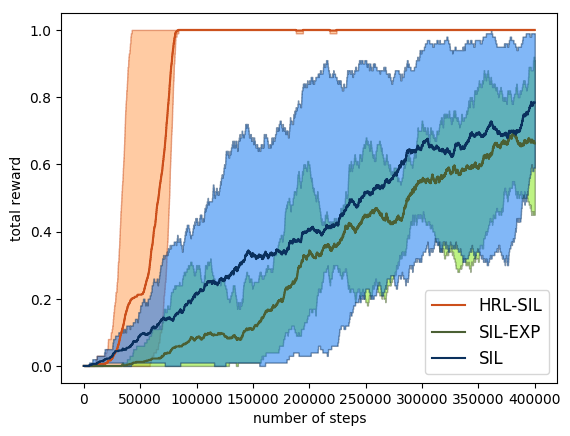} }}%
    \caption{ Results in Key-door-treasure-1.}%
    \label{fig:Exploration1}%
\end{figure}

In Key-door-treasure-1 (Figure~\ref{fig:Exploration1}) we observe that when the reward is given for every object, all the algorithms perform well, while by making the reward more sparse, our algorithm clearly outperforms the others, because of its ability to act on different timescales through the compressed state space and the option action space. 

\begin{figure}
	\includegraphics[width=0.25\textwidth]{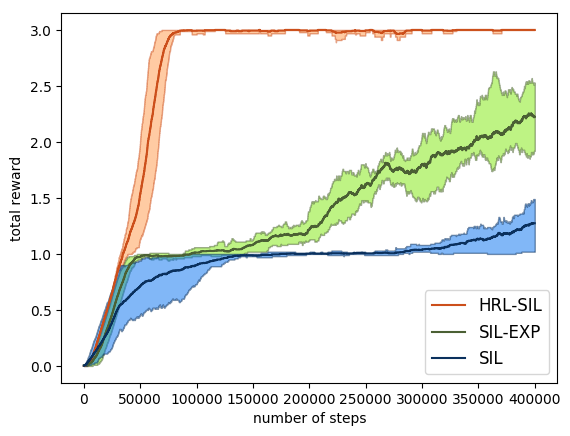}
	\centering
	\caption{Results in Key-door-treasure-2, reward for all objects.}
	\label{fig: Exploration2}
\end{figure}

We further investigate this in Key-door-treasure-2 (Figure~\ref{fig: Exploration2}) where the key and door are placed in different rooms. This makes exploration harder, and SIL struggles even in the setting with intermediate rewards, only learning to pick up the key, while SIL-EXP slowly learns to open the door and get the treasure thanks to the exploration bonus.

\subsection{Transfer Learning}\label{sec: Transfer Learning}

To investigate the transfer ability of the algorithm, we train 'HRL-SIL' subsequently on a set of tasks and compared to 'SIL-EXP'. In the first task, the goal is just to pick up a key and open a door. Once trained on this task, the agent is presented with a more complex task that also involves a treasure. The third task is the same as the second with the location of the objects mirrored.

\begin{figure*}[t]
    \centering
    \qquad
    \qquad
    \subfloat[Env 0]{{\includegraphics[width=3cm]{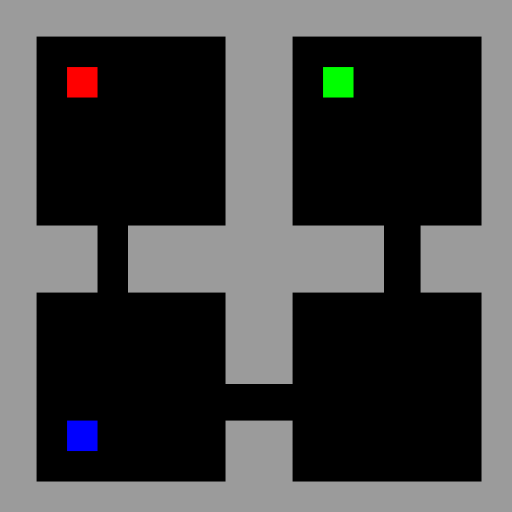} }}%
    \qquad
    \qquad
    \subfloat[Env 1]{{\includegraphics[width=3cm]{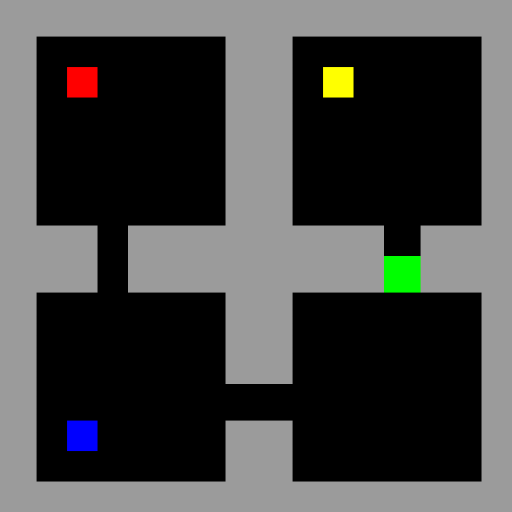} }}%
    \qquad
    \qquad
    \subfloat[Env 2]{{\includegraphics[width=3cm]{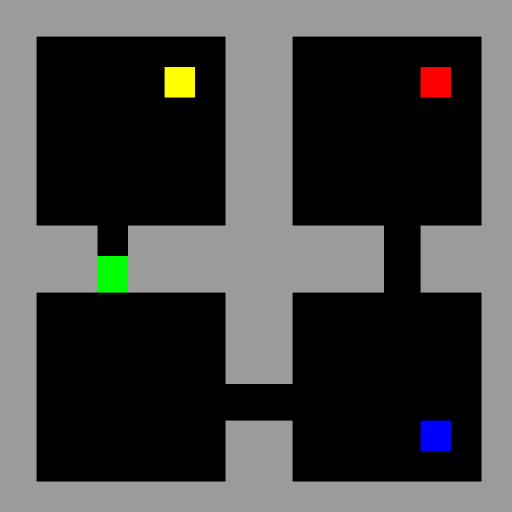} }}%
    \newline
    \subfloat[results in Env 0]{{\includegraphics[width=4cm]{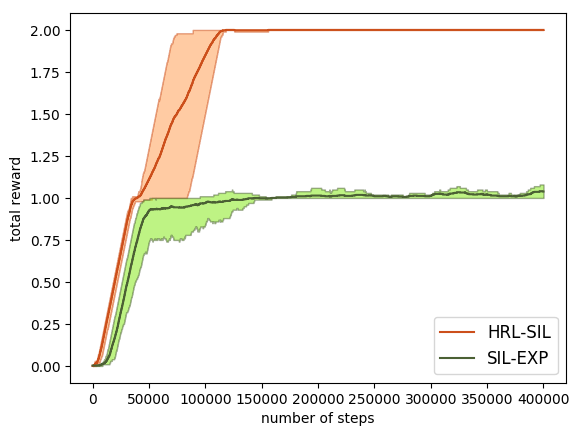} }}%
    \qquad
    \subfloat[results in Env 1]{{\includegraphics[width=4cm]{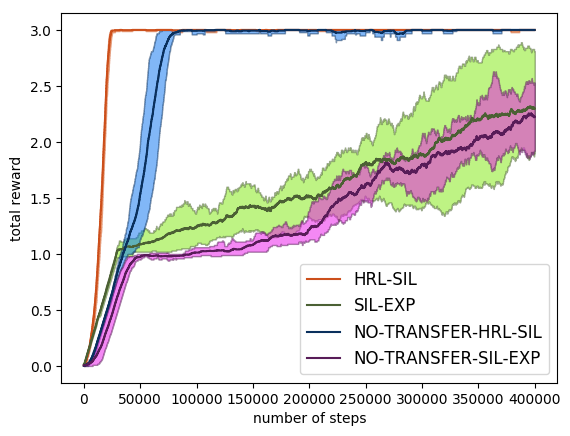} }}%
    \qquad
    \subfloat[results in Env 2]{{\includegraphics[width=4cm]{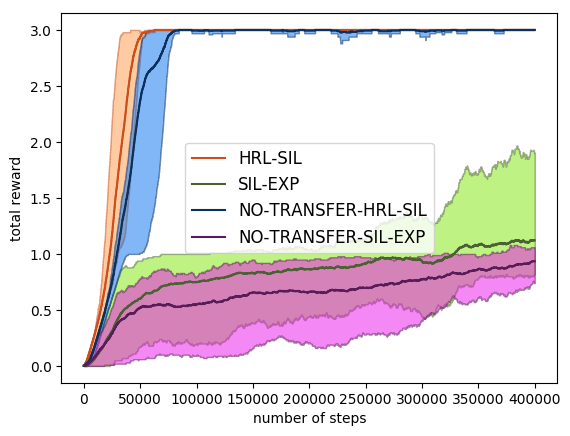} }}%
    \caption{ Results of transfer learning, with reward given for all objects.}%
    \label{fig: Transfer Learning}%
\end{figure*}

Our agent is evaluated by resetting the manager policy from task to task, while 'SIL-EXP' is evaluated by clearing the Experience Replay buffer between every task. We omit 'SIL' since it always performs worse than 'SIL-EXP'. From Figure~\ref{fig: Transfer Learning} we observe that the learned set of options $O$ and set of regions $Z$ transfer well across tasks. In contrast, 'SIL-EXP' struggles to solve new tasks. In the figure, 'NO-TRANSFER-HRL-SIL' and 'NO-TRANSFER-SIL-EXP' refer to the versions that relearn tasks from scratch.

\subsection{Controllability}\label{sec: ResultControllability}

\begin{figure}
	\includegraphics[width=0.25\textwidth]{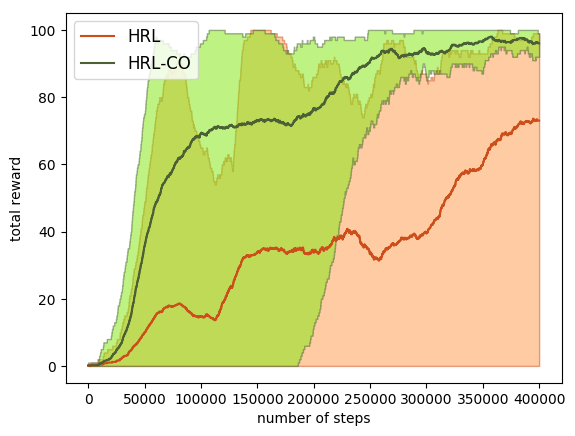}
	\centering
	\caption{Results in Montezuma's Revenge with controllability.}
	\label{fig: Montezuma}
\end{figure}

Lastly we test whether the controllability proxy helps transition successfully between regions. We compare two versions of our algorithm, one with controllability ('HRL-CO') and one without ('HRL'), in the first room of Montezuma's Revenge with the task of collecting the key. This environment is challenging, since the agent could learn unsafe transitions that lead to successful moves between regions but subsequently dying. As we can see from Figure~\ref{fig: Montezuma} the controllability proxy does indeed help in learning successful and safe transitions between regions, outperforming the simpler reward scheme of 'HRL'.

\begin{figure}[H]
\centering
$
\begin{array}{ll}
    \hline \text { Hyperparameters } & \text { Value } \\
    \hline \text { Architecture } & \text { -FC(64) } \\ & \text { -FC(64) } \\
    \text { Learning rate } & 0.0007 \\
    \text { Environments } & \text{Key-door-treasure} \\ & \text{MontezumaRevenge-}\\ & \text{-ramNoFrameskip-v4}\\
    \text { Number of steps per iteration } & 6 \\
    \text { Entropy regularization ( } \alpha \text { ) } & 0.01 \\
    \hline \text { SIL update per iteration ( } M \text { ) } & \mathrm{SIL}: 4, \mathrm{HRL}: [1,4]\\
    \text { SIL batch size } & 512 \\
    \text { SIL loss weight } & 1 \\
    \text { SIL value loss weight ( } \beta^{s} \text { il) } & 0.01 \\
    \text { Replay buffer size } & 10^{4} \\
    \text { Exponent for prioritization } & 0.6 \\
    \text { Bias correction, prioritized replay } & 0.4 \\
    \hline 
    \text{ Manager $\epsilon$-greedy} & [0.05, 0.005] \\
    \hline
    \text{ Count exploration $\beta$} & 0.2 \\
    \hline 
    \text{ Observation in Key-door-treasure} & (x, y, inventory) \\
    \text{ Observation in Montezuma} & (x, y) \\
\end{array}
$
\caption{ Hyperparameters used in the experiments.}%
\label{fig: Hyperparameters}%
\end{figure}

\section{Related work}\label{sec: related-work}

	Hierarchical reinforcement learning \cite{dayan1993feudal,sutton1999between,dietterich2000hierarchical} has a long history. Of particular relevance to this work are algorithms that automatically discover goals \cite{florensa2017automatic,bacon2017option,levine2020unsupervised}. The ability to compress the state space is also critical to our work \cite{mannor2004dynamic,vezhnevets2017feudal}. Design choices of how to use the task compression, and how to distribute the reward, identifies different instances of such methods.

	Our compression function is similar to that of Go-Explore \cite{ecoffet2019go}, which also partitions the state space into regions and performs greedy best-first search to solve Montezuma's revenge. The main difference is that our algorithm can learn near-optimal policies for transitioning between regions, while Go-Explore does not improve on the first action sequence generated randomly.

	Other authors have proposed algorithms for sparse-reward domains that involve a notion of hierarchy. \citet{keramati2018learning} propose a model-based framework to solve sparse-reward domains, and incorporate macro-actions in the form of fixed action sequences that can be selected as a single decision. \citet{shang2019learning} use variational inference to construct a world graph similar to our region space. However, unlike our model-free method, the option policies are trained using dynamic programming, which requires knowledge of the environment dynamics. \citet{eysenbach2019search} build distance estimates between pairs of states, and use the distance estimate to condition reinforcement learning in order to reach specific goals, which is similar to defining temporally extended actions.

\section{Discussion}\label{sec: discussion}

In spite of the encouraging results in Section~\ref{sec: experiments}, the current version of the proposed algorithm has several limitations. In this section we discuss potential future improvements aimed at addressing these limitations.

\paragraph{Invariant state-action space} The current version of the algorithm assumes that the agent has prior knowledge of the invariant part of the state-action space, i.e.~$S_i\times A_i$. In some applications, this seems like a reasonable assumption, e.g.~in environments such as MineCraft or DeepMind Lab where the agent has access to a basic set of actions, and is later asked to solve specific tasks. In case prior knowledge of $S_i\times A_i$ is not available, previous work on lifelong learning has shown how to automatically learn a latent state space that is common to a range of tasks \cite{bouammar2015lifelong}.

\paragraph{Compression function} The algorithm also assumes that the agent has access to a compression function $f$ which maps invariant states to regions. In case such a function is not available, the agent would need to automatically group states into regions. We believe that the algorithm is reasonably robust to changes in the compression function, but an important feature is that neighboring states should be grouped into the same region. Dilated recurrent neural networks \cite{chang2017dilated} are designed to maintain constant information during a given time period, similar to the idea of remaining in a given region for multiple timesteps, and have been previously applied to hierarchical reinforcement learning \cite{vezhnevets2017feudal}.

\paragraph{Option policies} Another limitation of the algorithm is that it needs to learn a large number of policies which scales as the number of regions times the number of neighbors. In large-scale experiments it would be necessary to compress the number of policies in some way. Since regions are mutually exclusive, in principle one could use a single neural network to represent the policy of $|Z|$ different options. However, in preliminary experiments such a representation suffers from catastrophic forgetting, struggling to maintain the optimal policy of a given option while training the policies of other options. We believe that a more intelligent compression scheme would be necessary for the algorithm to scale, potentially sharing a single policy among a carefully selected subset of options.

\section{Conclusion}\label{sec: conclusion}

We presented a hierarchical reinforcement learning algorithm that decomposes the state space using a compression function and introduces subtasks that consist in moving between the resulting partitions. We illustrated that the algorithm can successfully solve relatively complex sparse-reward domains. As discussed in Section~\ref{sec: discussion}, there are many opportunities for extending the work in the future.

\newpage\phantom{blabla}
\newpage\phantom{blabla}

\bibliography{reference.bib}
\bibliographystyle{src/icml2020}

\end{document}